# Learning Representations from Deep Networks Using Mode Synthesizers

**N.E. Osegi [c,*], P. Enyindah [b]**

a. National Open University of Nigeria
b. University of Portharcourt, Rivers State, Nigeria

**Abstract**

Deep learning Networks play a crucial role in the evolution of a vast number of current machine learning models for solving a variety of real world non-trivial tasks. Such networks use big data which is generally unlabeled unsupervised and multi-layered requiring no form of supervision for training and learning data and has been used to successfully build automatic supervisory neural networks. However the question still remains how well the learned data represents interestingness, and their effectiveness i.e. efficiency in deep learning models or applications. If the output of a network of deep learning models can be beamed unto a scene of observables, we could learn the variational frequencies of these stacked networks in a parallel and distributive way. This paper seeks to discover and represent interesting patterns in an efficient and less complex way by incorporating the concept of Mode synthesizers in the deep learning process models.

1. **Introduction**

In the world of deep learning several important contributions have been made with good successes. Hinton et al [1] introduced the concept of Restricted Boltzmann Machines (RBM's) which have been successfully applied to image [2] and speech [3, 4] recognition tasks and more recently to multi-modal learning [5, 6] and video [7] recognition tasks with some modifications in architecture, such as addition of Rectified Linear Units (ReLU's) [8, 9] and drop-outs [10]. Auto-Encoders, a versatile and automatic way of encoding data into meta-patterns without the cost implications of RBM's have been proposed and developed fully by several researchers including but not limited to Ordinary Auto-Encoder (OAE), Sparse Auto-Encoders (SAE) and De-noising Auto-Encoders (DAE) [11]. Also, the resurgence of the Long-Short-Term-Memory (LSTM) introduced in [13] is gradually gaining grounds in a variety of applications such as in speech processing [14] and video representations [15]. However as stated in [11], there still remain a large set of unanswered questions in the deep learning field and thus direct learning systems are desired [12]. One main challenge of deep learning networks is learning representations of fast changing or random data with a Gaussian unit. In this study, mode synthesizers will be introduced as a candidate solution for learning variational data i.e. scenes with fast changing but probably reoccurring data units. The system is adaptable and can be modified by any deep neural network researcher. In Section 2, we introduce the Mode synthesizer's concept. Section 3 presents some experiments using subset of benchmark dataset. We conclude the paper in Section 4.

*Corresponding author. Tel.: +234 703-008-1615.
E-mail addresses: nd@osegi.com (N.E. Osegi)

## 2. Mode Synthesizers (MoS)

The MoS is a set of routines that encodes frequencies with the goal of reconstructing an approximate replica of the input for learning. Basically, an input data stream set is fed to the MoS System and it performs a reconstruction similar to that of an Auto-Encoder but with a bias for computing the modes of the sequence data. In this section, we briefly introduce the MoS system and give some theoretical insights into its operation.

### 2.1 MoS Architecture

A mode synthesizer (MoS) typically encodes frequencies from an unsupervised system or network to learn a representation. The system may be as simple as a single RBM or OAE to more complex versions or Multi-Layer Networks. Once a representation is learned, the MoS can then proceed with discrimination tasks just like in RBM'S or Auto-Encoders. Figure 1 shows the architectural model of a MoS System which may employ any standard representation of a Multi-Layer Network for supervised training. The system is well suited for learning tasks that are distributive, multi-modal and in which sequences of data reoccur over time. This is common scenario in real world situations. Figure 2 shows the building block for implementing MoS with a simple architecture. More units may be stacked together to generate more complex reconstructions.

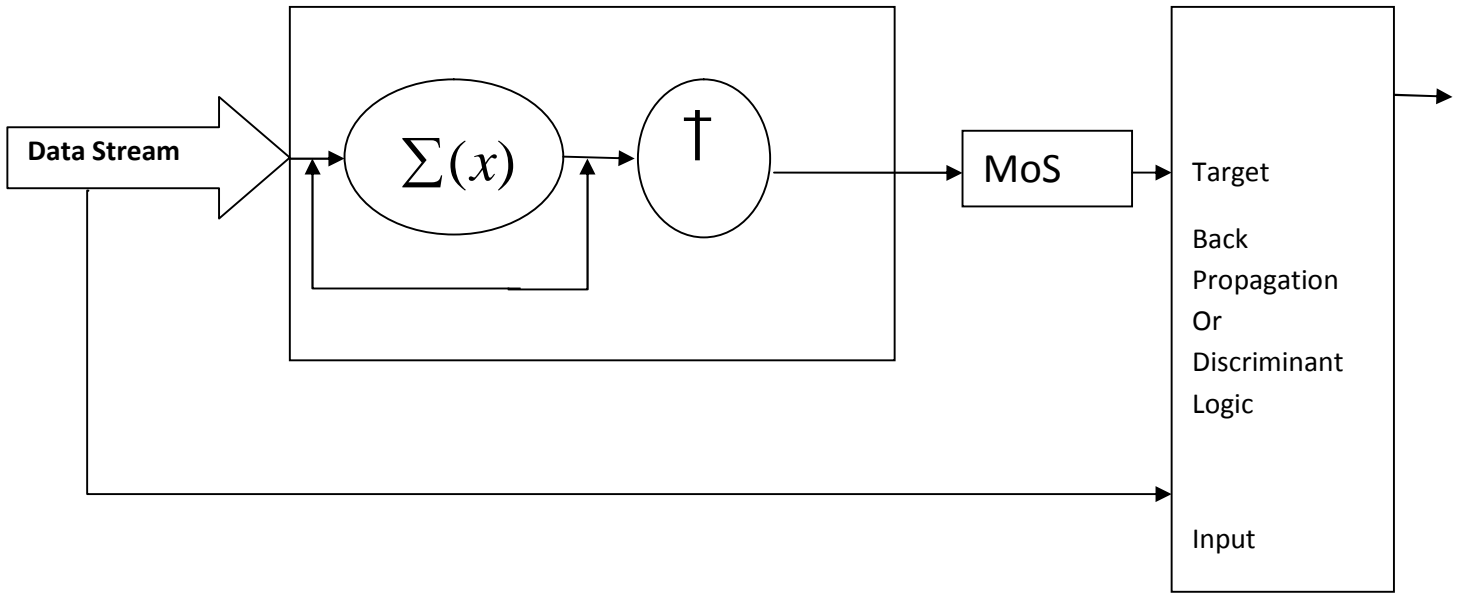

Fig1. Architectural Design for a MoS System

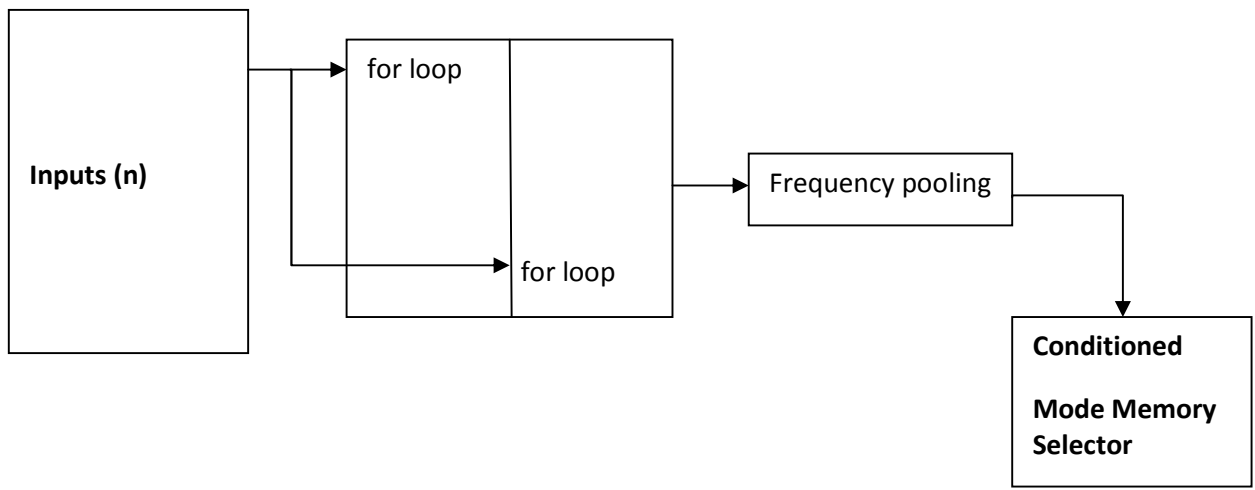

Fig2. MoS Building Block

## 2.2 Analytical Formulation of a MoS System

Figure 1 shows the proposed architectural diagram of a MoS system. Here, the goal is to perform a reconstruction of the input using mode synthesizer as a learning unit for the output activations $_i$, apply these reconstructions as a target for the original input system, then feed to a suitable back-propagation network. Mathematically speak, we define a reconstruction as in an Auto-Encoder [16]:

$$Z(x) = \otimes(h(x))\{ \begin{matrix} w_i = 1 \\ bi_i \cong 0 \end{matrix} \quad (1)$$

For which we may seek to achieve an approximate minimization of the function:

$$f_{min}(x) = \arg\min\_approx \frac{1}{N}\sum_{i=1}^{N} L_\succ(x^{(i)}, z(x^{(i)}))$$

Where wi = weights of input units,

bi = biases

h = hidden unit activations

x = input units $x \in \Re^+$

$\otimes$ = activation function, which could be any of logistic sigmoid, linear, elliot, hyperbolic tangent functions.

z = output activations or reconstructions

$L_\succ$ = conditional relation

We then perform the following steps to compute the output activations:

$$Z_k = [Z_k, \sum\sum(\{x_i\})] \quad (2)$$

$$Z_k^+ = \sum\sum Z_k \quad (3)$$

$$Z_{kp} = Z_k(x)./Z_k^+ \quad (3)$$

$$f_v = Z_{kp} + bi \quad (4)$$

$$t_{f_i} = \dagger(f_v) \tag{5}$$

Where, $Z_k$ is the recurrent summation of the second-order of the inputs,

$Z_k^+$ = net sum of $Z_k$

$Z_{kp}$ = deterministic probabilities inferred

$f_v$ = hidden activations and,

$t_f$ = transfer function of the output activations

### 2.2.1 Mode-Learning Rules

We synthesize mode units using a conditioned based approach by enforcing the following rules:

i.   Maximal Rule – if $Z_{kp} >$ ts1
ii.  Minimal Rule - if $Z_{kp} <$ ts2
iii. Equivalency Rule - if $Z_{kp} == 1$
iv.  Fuzzy Rule – if $Z_{kp} > 0.5$ Or $Z_{kp} < 1$

Where ts1, and ts2 are threshold points between 0 and 1.

Mode pooling is achieved by calling mode memory unit's functionality as desired and passing onto the back-propagation stage. In this way we encourage diversity in the mode selection process. These rules may also be seen as enforcing conditional penalties for constraining the MoS from deviating significantly from an expected range.

### 3. Experiments and Results

Simulation experiments have been conducted on a single-core processor with a clock speed of 2.0GHz, 160GB main memory and a 2.0GB RAM. The program is developed as an M-file function in Matlab version 7.5 and can be accessed from [19]. We have used a subset of the MNIST database of handwritten digits from [17] which can also be accessed in its reduced form from [18. Sachin cis.jhu.edu]. Pre-processing instructions is included in the M-file program and the program may be run fro a user generated M-file.

The following operations may be performed:

i.   Set the number of generations to 3. Higher generations may not necessarily result in better results

    ii.    Change the number of digits sampled "n". This has the effect of increasing the reconstruction bandwidth. Since no batches are employed, this will reduce to a single batch operation.

    iii.    Change the 'digit' class. We have a maximum of 10 digits from 0 to 9.

These three parameters may be tuned to study the model performance.

**Sampled Instances**

Simulation run for n = 12 and class 'digit0', 'digit1' and digit8' is as shown in Figure 3 to Figure 5. The larger than number size should be attributed to its recurrent structure and frequency duplication principle enforced in the architecture.

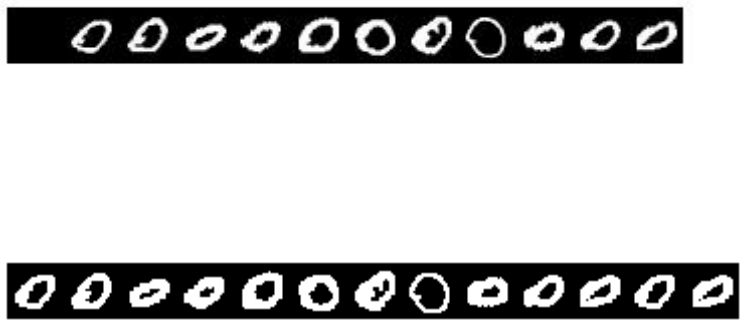

Figure3. Original and Reconstructed Image Using Mode-Synthesizer Algorithm for digit 0

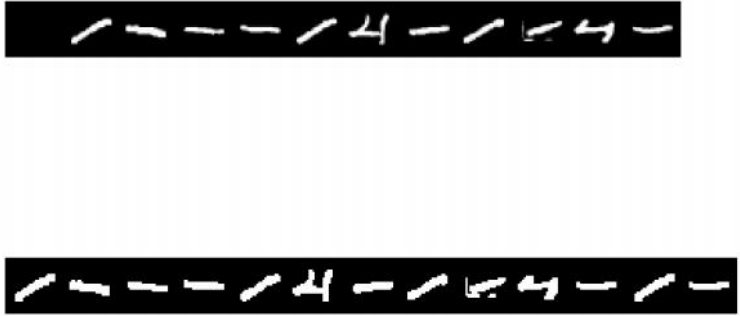

Figure4. Original and Reconstructed Image Using Mode-Synthesizer Algorithm for digit 1

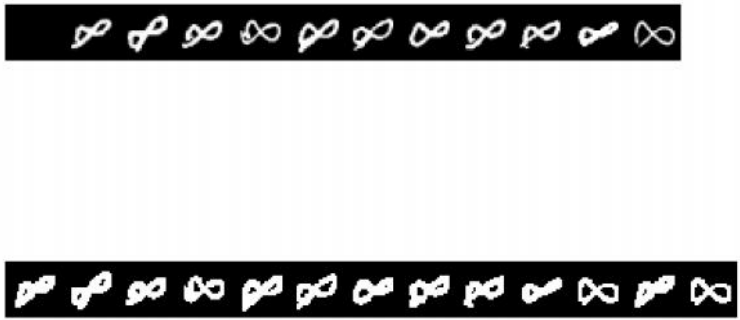

Figure5. Original and Reconstructed Image Using Mode-Synthesizer Algorithm for digit 8

Simulation run for class 'digit0' at n = 50 is as shown in Fig6. This shows as expected a much larger bandwidth. Table1 gives the learned indices of the different digit classes at n = 100. This clearly shows some variation in digits 2 and 8 while the rest have similar learned indices. There is every reason to believe that variation will reoccur at higher samples much greater than a hundred, however, we limit ourselves to the visible range.

Table1. Learned Indices for Different Handwriting digit classes at n = 100

| s/n | Digit 0 | Digit 1 | Digit 2 | Digit 3 | Digit 4 | Digit 5 | Digit 6 | Digit 7 | Digit 8 | Digit 9 |
|---|---|---|---|---|---|---|---|---|---|---|
| 1 | 1 | 1 | 20 | 1 | 1 | 1 | 1 | 1 | 48 | 1 |
| 2 | 99 | 99 | 58 | 99 | 99 | 99 | 99 | 99 | 76 | 99 |
| 3 | 100 | 100 | 97 | 100 | 100 | 100 | 100 | 100 | 98 | 100 |
| 4 | none | none | 99 | none | none | none | none | none | 100 | none |
|   |   |   |   |   |   |   |   |   |   |   |

## 4. Conclusions

Mode synthesizers (MoS) are a simple yet effective approach for learning deep generative models. The features learned are purely statistical using the frequency of reoccurring data sequences and may further be incorporated into existing deep learning models and architectures to improve discrimination tasks. Simply put, MoS learns the most frequent data sequences first, and then learns the less frequent ones in subsequent iteration. This builds a balance between shallow learning and deep learning. Experiments, suggest that MoS should perform well for interesting data i.e. when the data sequences are regular or consists of repetitive features and a little less better when data is random. In a real world scenario, this can be achieved by temporal-recurrent restructuring. Although we have introduced some ideas for MoS, we believe there are better explanations and theories that describe this process. For instance, the MoS could be designed to pool the minimum set when the data is largely irregular and random about the Gaussian. MoS should be promising as well for constrained devices, particularly in embedded applications.